\pdfoutput=1
\documentclass[11pt]{article}

\usepackage[]{ACL2023}

\usepackage{times}
\usepackage{latexsym}

\usepackage[T1]{fontenc}
\usepackage[utf8]{inputenc}

\usepackage{microtype}

\usepackage{inconsolata}

\usepackage{graphicx}
\usepackage{multirow}
\usepackage{colortbl}
\usepackage{amsmath}
\usepackage{float}
\usepackage{enumitem}
\usepackage{url}

\title{FANTAstic SEquences and Where to Find Them: \\Faithful and Efficient API Call Generation \\through State-tracked Constrained Decoding and Reranking}

\author{Zhuoer Wang$^{\dagger1}$\quad Leonardo F. R. Ribeiro$^2$ \\
\textbf{Alexandros Papangelis$^2$ \quad Rohan Mukherjee$^2$ \quad Tzu-Yen Wang$^2$} \\
\textbf{Xinyan Zhao$^2$\quad Arijit Biswas$^2$\quad James Caverlee$^1$\quad Angeliki Metallinou$^2$} \\
$^1$Texas A\&M University \quad $^2$Amazon\\
\texttt{wang@tamu.edu}
}

\begin{document}
\maketitle

\renewcommand{\arraystretch}{1.2}

\renewcommand{\thefootnote}{\fnsymbol{footnote}}
\footnotetext[2]{The major portion of the research was done during an internship at Amazon.}
\renewcommand{\thefootnote}{\arabic{footnote}}

\begin{abstract}
API call generation is the cornerstone of large language models' tool-using ability that provides access to the larger world. However, existing supervised and in-context learning approaches suffer from high training costs, poor data efficiency, and generated API calls that can be unfaithful to the API documentation and the user's request. To address these limitations, we propose an output-side optimization approach called FANTASE. Two of the unique contributions of FANTASE are its State-Tracked Constrained Decoding (SCD) and Reranking components. SCD dynamically incorporates appropriate API constraints in the form of Token Search Trie for efficient and guaranteed generation faithfulness with respect to the API documentation. The Reranking component efficiently brings in the supervised signal by leveraging a lightweight model as the discriminator to rerank the beam-searched candidate generations of the large language model. We demonstrate the superior performance of FANTASE in API call generation accuracy, inference efficiency, and context efficiency with DSTC8 and API Bank datasets. 
\end{abstract}
\section{Introduction}

\begin{figure}[t]
    \centering
    \includegraphics[width=0.44\textwidth]{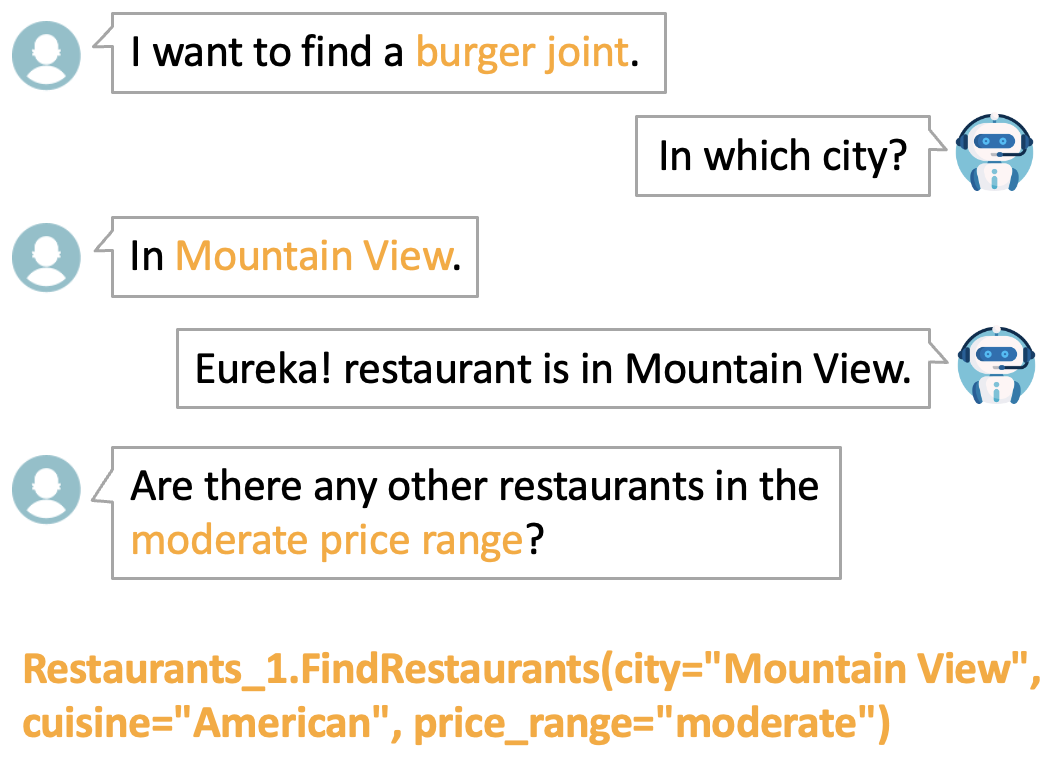}
    \vspace{-6pt}
    \caption{Example of an API call that retrieves information based on the user's needs given in the conversation.}
    \vspace{-15pt}
    \label{figure_intro}
\end{figure}

In recent year, there has been a surge of interest in enabling the automated tool-using capability of intelligent systems~\cite{Schick2023ToolformerLM, mialon2023augmented}. 
Specifically, as a bridge to the larger world, Application Programming Interface (API) calls allow virtual assistants to control smart-home devices, retrieve information, make reservations, and more on the user's behalf. 
Figure~\ref{figure_intro} shows how an API call may improve the user-assistant conversation and satisfy the user's needs. Generating such an API call requires advanced capabilities in understanding the requirements of an API (including its endpoints, parameters, and expected data formats) and reasoning over the conversation context to translate the user's needs into the appropriate API format. 

With recent breakthroughs in generative Large Language Models (LLMs) such as GPT-X~\cite{chatgpt-DBLP:conf/nips/Ouyang0JAWMZASR22,gpt4-DBLP:journals/corr/abs-2303-08774} and LLaMA~\cite{touvron2023llama,touvron2023llama2}, 
researchers have started to investigate their competence in complex reasoning tasks such as utilizing appropriate API tools~\cite{li2023apibank,qin2023toolllm, wang-etal-2023-measuring}. 
Their attempts focus on methods that can generally be grouped into those based on \emph{supervised fine-tuning} for task-specific usage and those based on augmenting input-side context information (such as API shortlisting and exemplar selection) and optimizing prompts for \emph{in-context learning}~\cite{NEURIPS2020_1457c0d6,NEURIPS2022_9d560961}. Despite the strong supervision or extensive context, these methods still cannot ensure the generation's faithfulness with respect to the API documentation and suffer data and compute inefficiency.
In contrast to previous works, we focus on how decoding strategies improve the generation's faithfulness, which is complementary to supervised fine-tuning and in-context learning methods. As a result, we present FANTASE (FANTAstic SEquences and Where to Find Them), a framework that employs State-tracked Constrained Decoding (SCD) and Reranking components, for faithful and efficient API call generation.

The SCD tracks the states of the generation and retrieves appropriate API documentation constraints in the form of Constrained Token Search Trie (CTST) used at each decoding step. SCD is guaranteed to generate API calls that are faithful with respect to the API documentation (\S~\ref{sec:scd}), and provides inference efficiency (\S~\ref{sec:results-inf_eff}) with CTST that eliminates unnecessary forward inference passes. 
Compared to supervised fine-tuning methods, SCD brings considerable improvements (\S~\ref{sec:results-gen_acc}) without the data labeling and model training related hefty costs of labor, time, and computing that become increasingly expensive as the size of the LLM grows~\cite{yang2023harnessing}. 
SCD also reduces the in-context learning's reliance on the repeated supply of extensive contextual information for the inference of each instance (\S~\ref{sec:results-con_eff}) by effectively incorporating API documentation constraints and guaranteeing the associated faithfulness at the decoding stage. 

The Reranking component of FANTASE leverages models that are significantly smaller than LLMs for efficient incorporation of supervised signals (\S~\ref{sec:reranking}). As the correct API generation may not always have the highest sequence probability among beam-searched candidate sequences (\S~\ref{prelim}), we train lightweight models to discriminate and rerank LLMs' candidate generations and demonstrate their effectiveness in digging out those correct sequences (\S~\ref{sec:results-gen_acc}). 
Compared to the supervised fine-tuning of LLMs, the Reranking component features extremely low training costs as it employs lightweight models. Compared to input-side optimized in-context learning methods, the Reranking component can address the severe performance issue associated with the absence of valuable supervised signals. 
Notably, FANTASE is a approach that suits the evolving and vast nature of real-world APIs. With the update of API documentation or the application to the new domain, LLMs fine-tuned with old data would require re-tuning with new data~\cite{kumar2022finetuning}. For FANTASE, SCD can adapt by constraining the decoding with a new set of constraints elicited from the new API documentation, while re-tuning the lightweight Reranking models has lower time and compute cost.

\noindent In summary, we make the following novel contributions:
\begin{itemize}[noitemsep, topsep=0pt, partopsep=0pt]
    \item We propose State-tracked Constrained Decoding that can effectively enforce constraints elicited from API Documentation, which yields faithful generation and context efficiency. % and better resilience to API evolution
    \item We leverage Constrained Token Search Trie to reduce unnecessary forward inference passes, which yields faster generation speed.
    \item We demonstrate the effectiveness of incorporating supervised signals with a small model to discriminate and rerank the beam-searched candidate generations of LLMs. 
\end{itemize}

\begin{table*}[t]
\scriptsize
\centering
\begin{tabular}{|c|l|}
\hline
\textbf{\begin{tabular}[c]{@{}c@{}}Related\\ Conversation\end{tabular}}      & \begin{tabular}[c]{@{}l@{}}Human: I want to find a burger joint.      Assistant: In which city?.\\ Human: In Mountain View.                     Assistant: Eureka! restaurant is in Mountain View.\\ ...... Human: Are there any other restaurants in the moderate price range?\end{tabular}                                                                                                                                                         \\ \hline
\textbf{\begin{tabular}[c]{@{}c@{}}Related\\ API Documentation\end{tabular}} & \begin{tabular}[c]{@{}l@{}}...... Restaurants\_1.FindRestaurants("cuisine" : Required, "city" : Required, "price\_range" : Optional, \\ "has\_live\_music" : Optional, "serves\_alcohol" : Optional) ...... the possible values for "cuisine" include \\ {[}"Mexican", "Chinese", "Indian", "American", "Italian"{]} ......\end{tabular}                                                                                                             \\ \hline
\textbf{Expected API Call}                                                   & \texttt{Restaurants\_1.FindRestaurants(city="Mountain View", cuisine="American", price\_range="moderate")}                                                                                                                                                                                                                                                                                                                                                    \\ \hline
\textbf{Top Candidates}                                                      & \begin{tabular}[c]{@{}l@{}}1. \texttt{Restaurants\_1.FindRestaurants(price\_range="moderate", city="MountainView")} \textcolor{red}{missing \texttt{cuisine}}\\ 2. \texttt{Restaurants\_1.FindRestaurants(cuisine="\textcolor{red}{Burgers}", city="MountainView")} \textcolor{red}{missing \texttt{price\_range}}\\ 3. \texttt{Restaurants\_1.FindRestaurants(cuisine="American", city="MountainView")} \textcolor{red}{missing \texttt{price\_range}}\\ 4. \texttt{Restaurants\_1.FindRestaurants(price\_range="moderate")} \textcolor{red}{missing \texttt{cuisine} and \texttt{city}}\\ 5. \texttt{Restaurants\_1.FindRestaurants(cuisine="American", city="MountainView", price\_range="moderate")}\end{tabular} \\ \hline
\end{tabular}
\caption{Preliminary analysis sample. With regular beam search decoding, the correct generation is only ranked the 5th, and other higher ranked generations exhibit various errors highlighted in \textcolor{red}{red}.}
\vspace{-11pt}
\label{table1}
\end{table*}

\section{Related Work}
\label{related_work}
\noindent\textbf{Constrained Decoding} offers controllable text generation by enforcing certain constraints at the decoding stage. Early research \cite{hokamp-liu-2017-lexically,post-vilar-2018-fast} concentrated on lexical constraints that enforce the inclusion of specific words or phrases in the outputs, which often neglects broader syntactic or semantic relationships. Later on, \citealt{lu-etal-2021-neurologic} introduced \emph{NeuroLogic Decoding} that handles more complex lexical constraints expressed by predicate logic. The subsequent extension, \emph{NeuroLogic A*esque Decoding}~\cite{lu-etal-2022-neurologic}, incorporated a lookahead heuristic to estimate future lexical constraint satisfaction. More recently, \citealt{NEURIPS2022_ab63a1a3} and \citealt{bastan-etal-2023-neurostructural} proposed parsing-based constrained decoding algorithms that tackle the challenge of ensuring correct syntactic relationships between word pairs. 

Specific to structured text generation, \citealt{scholak-etal-2021-picard} targeted Text-to-SQL generation and introduced \emph{PICARD} that checks the validity at each decoding step for SQL lexical and grammar correctness with incremental parsing. The latest advancement was made by \citealt{geng-etal-2023-grammar} who demonstrated that an incremental parser can be used with formal grammar on a much wider range of structured NLP tasks without finetuning. While the results are encouraging, existing methods require post-hoc constraint satisfaction checking or rely on dependency parsing at inference time, or both, which compromises the efficiency. The most recent and closest work to ours is API-aware Constrained Decoding~\cite{wang-etal-2023-measuring} that imposes function and argument token constraints based on API documentation. However, despite limited improvements, its decoding strategy results in a 20\% slowdown of the generation. In contrast to aforementioned methods, we achieve faster generation speed and guaranteed faithfulness with a novel State-tracked Constrained Decoding approach that dynamically incorporates appropriate constraints in the form of a retrieved token search trie.

\noindent\textbf{Discriminator Guided Generation} utilizes small discriminative models or external tools to guide the generation of LLMs. \citealt{DBLP:conf/iclr/DathathriMLHFMY20} proposed the Plug and Play Language Model concept that guides the generation of pretrained models with a lightweight attribute classifiers' gradient. However, it increases compute costs due to the extra forward and backward passes required for sampling and using the gradients from the attribute classifiers to push the pretrained model's hidden activations. Following works including \emph{GeDi}~\cite{krause-etal-2021-gedi-generative}, \emph{FUDGE}~\cite{yang-klein-2021-fudge}, and~\emph{BeamR} \cite{landsman-etal-2022-beamr} used different lightweight discriminators that classify the attribute of possible next tokens or partial sequence and reweigh token-level or beam-level probabilities at each decoding step towards the desired direction of attributes like sentiment, topic, formality, and so on. More recently, \citealt{ni2023lever} leveraged the execution results of a SQL executor to steer SQL generation, which achieved new state-of-the-art results. Nevertheless, the method is bounded by the prerequisite of the external executor. In our work, we employ a lightweight model to discriminate API call generation by the given context and perform a one-pass reranking of the beam-searched results, which brings in supervised signals effectively with little compute and time costs to the overall generation framework.

\begin{figure*}
    \centering
    \includegraphics[width=0.9\textwidth]{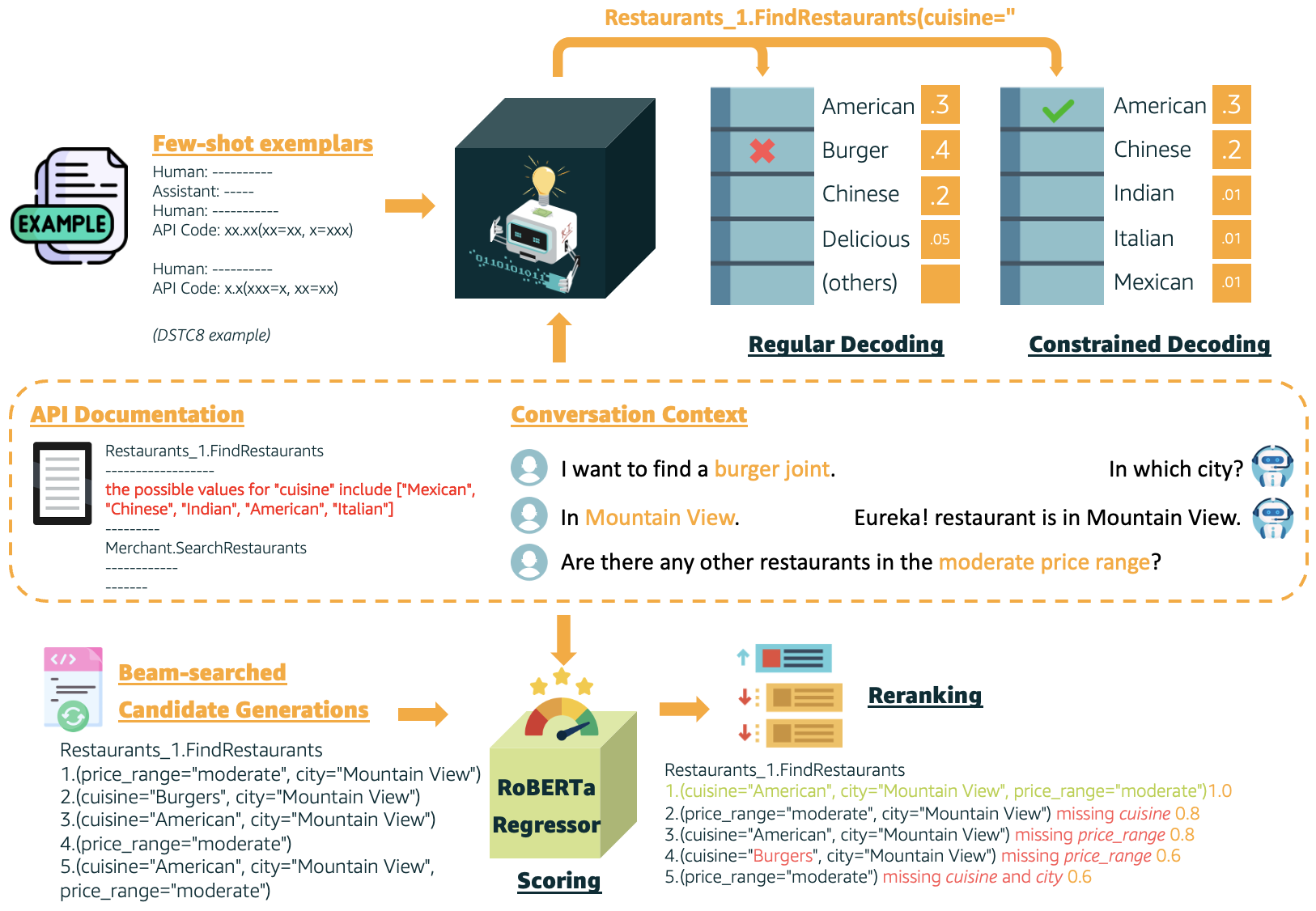}
    \caption{Illustration of the Concepts of Constrained Decoding and Reranking. (Upper half) Constrained Decoding enforces API documentation constraints and would only consider the five possible values of \texttt{cuisine}. (Lower half) A lightweight \texttt{RoBERTa} model is used to discriminate and rerank the beam searched candidate generations. }
    \vspace{-11pt}
    \label{figure_fantase}
\end{figure*}
\section{Preliminary Analysis}
\label{prelim}

To better understand the capabilities and limitations of existing LLMs on the task of API call generation, we conduct a preliminary inference analysis on one hundred DSTC8~\cite{dstc8kim}\footnote{Details will be given in Section~\ref{sec:data}} samples with an Alpaca~\cite{alpaca} model
that had been tuned with GPT-generated self-instruct~\cite{wang-etal-2023-self-instruct} data for better instruction following and in-context learning capabilities. We prompt the model with the DSTC8 data that contains task instruction, documentation of related APIs, two related exemplars, and conversation history. We use beam search with beam size 10 as the decoding algorithm, and we consider the top-10 high probability sequences as the candidate generations. 

Our quantitative analysis shows that for 73\% of the cases, the correct API calls are generated within those high probability sequences. However, within these cases, almost half of the correct sequences were not ranked as the highest, which yields a top-1 API call generation accuracy of 41\%. Table~\ref{table1} presents an example where the user wants to find a burger joint with a moderate price range in Mountain View. The supplied API documentation specified that the \texttt{Restaurants\_1.FindRestaurants} function has cuisine and city as the required arguments, and the \texttt{cuisine} argument has five possible values. However, the correct sequence was only ranked the 5th for the given example. All the other 4 candidates that have higher sequence probabilities missed some required arguments, and the second one also wrongly generated \texttt{Burgers} instead of one of the five possible values for the argument \texttt{cuisine}. 
Note that the model demonstrates some reasoning capability that can correctly map \texttt{Burgers} into \texttt{American} as shown in the second and the fifth sequences. Nevertheless, the overall sequence probability favors the problematic generation of \texttt{Burgers}, which may be attributed to the explicit mention of the word in the given conversation history. 

We conduct a further qualitative analysis to categorize the error types and possible mitigation for these cases. For the highest-ranked error cases, we find 33\% argument value error, 24\% missing required arguments, 19\% missing optional arguments, 14\% hallucination, and 10\% argument name error. 
Furthermore, 42\% of the error cases can be mitigated by enforcing the constraints described in the API documentation, 29\% of the cases require the better understanding of the conversation, and the remaining 29\% of the cases need a combination of the aforementioned two improvements. 

\section{The FANTASE Framework}

In Section~\ref{prelim}, we demonstrate that there are "FANTAstic SEquences"
 in the beam-searched candidate generations, and the question is where to find them. 
To dig out those "FANTAstic SEquences" with the data efficiency and compute efficiency in mind, we propose the FANTASE framework that consists of two major components -- State-tracked Constrained Decoding (\S\ref{sec:scd}) and Reranking (\S\ref{sec:reranking}), which aims at enforcing API constraints with guaranteed faithfulness to the API documentation and incorporating supervised signals at low compute costs respectively. 

\subsection{State-tracked Constrained Decoding}
\label{sec:scd}

In Figure~\ref{figure_fantase}, we illustrate the concept of State-tracked Constrained Decoding (SCD). For a regular decoding step, the consideration of the entire vocabulary space would lead to the high probability of the word \texttt{Burgers} overshadowing the correct word of \texttt{American}. To ensure the faithfulness to the API documentation, our SCD approach enforces the model to only consider the probabilities of the five possible values as documented in the API documentation. 

\begin{figure*}
    \centering
    \includegraphics[width=0.8\textwidth]{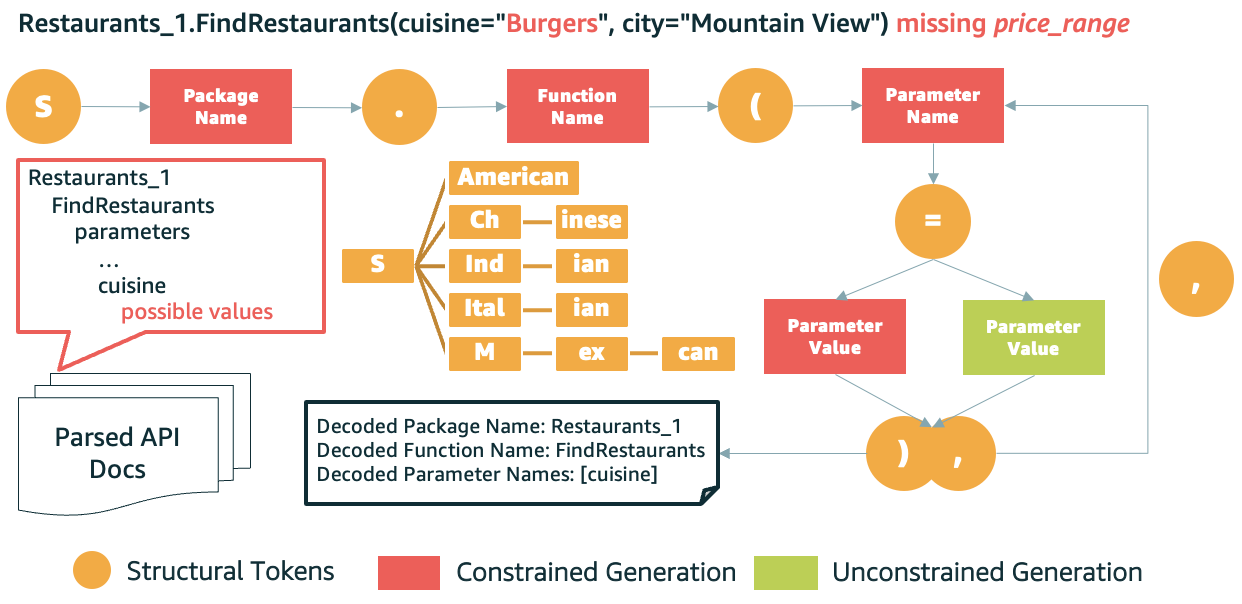}
    \vspace{-10pt}
    \caption{State-tracked Constrained Generation of API Call (showing the step of generating the value of parameter cuisine that has possible values of \texttt{American}, \texttt{Chinese}, \texttt{Indian}, \texttt{Italian}, and \texttt{Mexican}).}
    \vspace{-6pt}
    \label{figure_scd}
\end{figure*}

\begin{table*}[t]
\scriptsize
\centering
\begin{tabular}{|c|l|l|}
\hline
\textbf{Structural Token}                                  & \multicolumn{1}{c|}{\textbf{Generation State}}                                                            & \multicolumn{1}{c|}{\textbf{Actions}}                                                                                                                                                                                                                                                                                              \\ \hline
S (pseudo)                                                        & \multicolumn{1}{c|}{Start of the generation}                                                              & \begin{tabular}[c]{@{}l@{}}- Retrieve all package names\\ - Constrained generation of package name\end{tabular}                                                                                                                                                                                                                    \\ \hline
\begin{tabular}[c]{@{}c@{}}.\\ DOT\end{tabular}            & \multicolumn{1}{c|}{\begin{tabular}[c]{@{}c@{}}End of package name\\ Start of function name\end{tabular}} & \begin{tabular}[c]{@{}l@{}}- Record decoded package name\\ - Retrieve possible function names by using package name\\ - Constrained generation of function name\end{tabular}                                                                                                                                                       \\ \hline
\begin{tabular}[c]{@{}c@{}}(\\ LEFT\_BRACKET\end{tabular}  & \begin{tabular}[c]{@{}l@{}}End of function name\\ Start of argument name\end{tabular}                     & \begin{tabular}[c]{@{}l@{}}- Record decoded function name\\ - Retrieve possible argument names by using package and function names\\ - Constrained generation of argument name\end{tabular}                                                                                                                                        \\ \hline
\begin{tabular}[c]{@{}c@{}}=\\ EQUAL\end{tabular}          & \begin{tabular}[c]{@{}l@{}}End of argument name\\ Start of argument value\end{tabular}                    & \begin{tabular}[c]{@{}l@{}}- Record decoded argument name\\ - Retrieve possible argument values by using package, function, and argument names\\ - Check if the argument only takes certain possible values\\ -- If so, constrained generation of argument value\\ -- If not, perform regular unconstrained generation\end{tabular} \\ \hline
\begin{tabular}[c]{@{}c@{}},\\ COMMA\end{tabular}          & \begin{tabular}[c]{@{}l@{}}End of argument value\\ Start of argument name\end{tabular}                    & \begin{tabular}[c]{@{}l@{}}- Reuse previously retrieved possible argument names\\ - Constrained generation of argument name\end{tabular}                                                                                                                                                                                           \\ \hline
\begin{tabular}[c]{@{}c@{}})\\ RIGHT\_BRACKET\end{tabular} & End of the generation                                                                                     & \begin{tabular}[c]{@{}l@{}}- Check if the list of decoded argument name contains all the required arguments\\ -- If so, conclude the generation\\ -- If not, replace RIGHT\_BRACKET with COMMA and enforce continued generation\end{tabular}                                                                                       \\ \hline
\end{tabular}
\caption{State Tracking with Structural Tokens and Associated Actions.}
\vspace{-15pt}
\label{table2}
\end{table*}

Different from conventional token-occurrence-based constrained decoding approaches as described in Section~\ref{related_work}, our approach takes the relation between package, function, argument, and argument values into consideration. SCD allows precise and dynamic enforcement of constraints based on the API documentation and generated units, which avoids the look-ahead decoding and pruning as other constrained decoding algorithms would normally require. The implementation of SCD consists of three major parts: 1) extraction of constraints from API documentation, 2) state tracking of the generation for constraints retrieval, and 3) constrained decoding with token search trie. 

As API documentation is usually well-structured, it is feasible to extract constraints with simple rules. As a preprocessing step, we use regular expressions to extract five types of constraints including 1) available packages, 2) functions of each package, 3) required arguments of each function, 4) optional arguments of each function, and 5) possible values of each argument. We store these constraints in a lookup table with package name, function name, and argument name as the query keys. At the inference stage, we query the lookup table to fetch corresponding constraints by decoded package name, function name, and/or argument name. If the API is evolved with changing constraints such as new required/optional arguments, changing names, changing possible values, etc., new constraints can be enforced effortlessly at the inference stage by re-parsing the updated API documentation, which is less expensive than re-tuning the model with updated labeled data. 

In Figure~\ref{figure_scd}, we illustrate the SCD at the inference stage. To ensure appropriate constraints can be retrieved and enforced at the precise inference step of the generation, the state of the generation is determined by tracking the model-generated structural tokens. The structured nature of the API call results in signature tokens that indicate the end or start of different units of the API call as we specified in Table~\ref{table2}.
\begin{figure}
    \centering
    \includegraphics[width=0.5\textwidth]{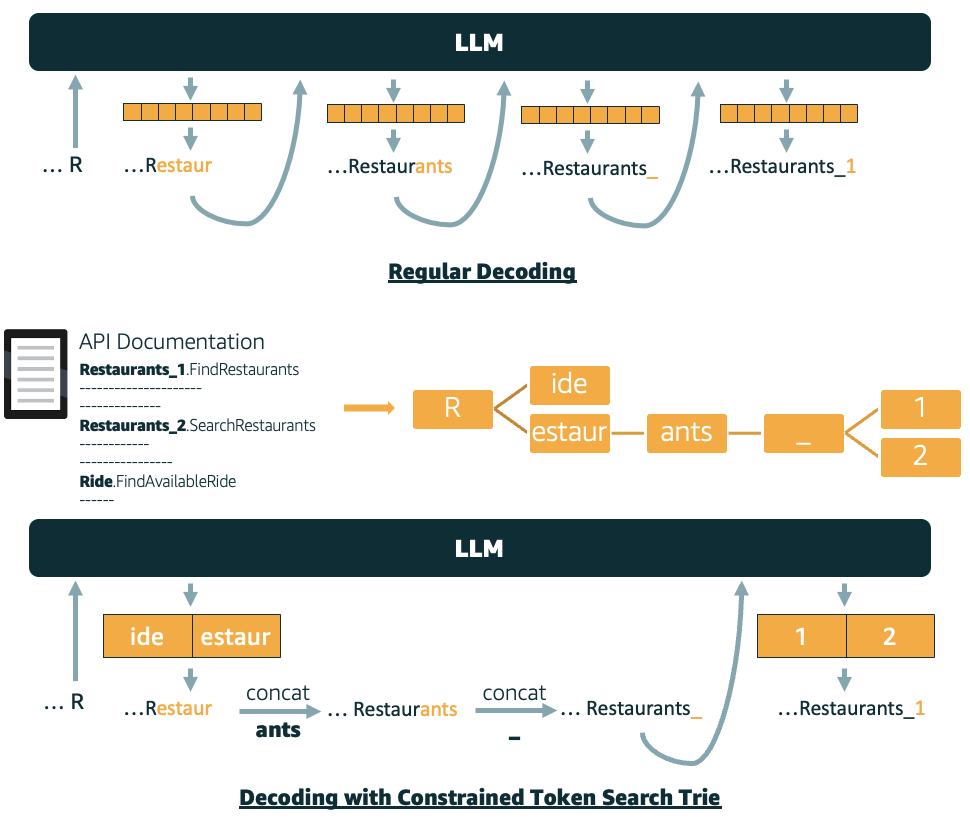}
    \vspace{-22pt}
    \caption{Comparison of the Generation of \texttt{Restaurants\_1} with Regular Decoding and Decoding with Constrained Token Search Trie. }
    \vspace{-11pt}
    \label{figure_ctst}
\end{figure}
Specifically, for the constrained generation of an API call unit, we enforce the model to decode along the Constrained Token Search Trie (CTST) as illustrated in Figure~\ref{figure_ctst}. The tokenizer of LLMs performs WordPiece tokenization, which breaks down the word into smaller subword tokens for various benefits~\cite{devlin-etal-2019-bert}. Accordingly, the package name \texttt{Restaurants\_1} would be autoregressively generated by the LLM piece by piece with five forward-pass inference steps as shown in the upper half of Figure~\ref{figure_ctst}. At the preprocessing step, we build the extracted constraints into CTST. When conducting constrained generation, the forward inference pass is only necessary for nodes that have multiple branches. In such a case, only the probabilities of the possible next tokens as indicated by the CTST would be considered. For nodes that only have one child, the subsequent token is directly appended, which saves the time and compute costs of a forward inference pass. 

For SCD, we implement it with four different sampling strategies including greedy search, top-k sampling, top-p sampling, and beam search.  

\subsection{Reranking}
\label{sec:reranking}
Although the SCD approach we introduced in Section~\ref{sec:scd} can guarantee the generated API call's faithfulness to the \emph{API documentation}, the faithfulness to the \emph{user's request} is solely dependent on the LLM's zero-shot or few-shot in-context learning and reasoning capabilities. Also, the valuable labeled training data hasn't been exploited yet with SCD. Fine-tuning the LLM might be a straightforward solution, but the huge compute costs associated with the growing parameter size of LLM motivates us to seek alternative solutions. 

Inspired by the studies of discriminator-guided generation, we propose the supervised training of a lightweight scorer for the reranking of beam-searched candidate generations. To train the scorer, we generate data as follows: we prompt the Alpaca-13B model with training set samples and obtain associated beam-searched candidate generations for each sample. For each candidate generation, the matching score with respect to the ground truth is calculated as the target of the scorer. Such data is used for the tuning of a RoBERTa-base\cite{roberta-DBLP:journals/corr/abs-1907-11692} model that has 125M parameters to predict the matching score based on the input of conversation and API Documentation context and the candidate generation. Specifically, we train the model with sample-wise batching that groups candidate generations of the same context into one mini-batch and use the MSE Loss and Spearman Soft Ranking Correlation Loss~\cite{10.5555/3524938.3525027} as the training objective for optimal performance.

Instead of returning the beam-searched candidate generation that has the highest sequence probability as the final generated API call, we use the trained scorer to discriminate each of the candidate generations and rerank them accordingly. This reranking strategy allows low-cost incorporation of task and domain-specific context reasoning capability learned from the valuable labeled data and complements the SCD approach and LLM's zero-shot or few-shot in-context learning and reasoning capabilities.  

\begin{table*}[h]
\scriptsize
\centering
\begin{tabular}{|ccc|}
\hline
\multicolumn{1}{|c|}{\textbf{Dataset}} & \multicolumn{1}{c|}{\textbf{\space\space DSTC8 (Two Exemplars) \space\space}} & \textbf{API Bank (No Exemplars)} \\ \hline
\multicolumn{3}{|c|}{\cellcolor[HTML]{EFEFEF}{\color[HTML]{000000} \textbf{In-context Learning Methods}}} \\ \hline
\multicolumn{3}{|c|}{\textbf{Baselines}}                                                                  \\ \hline
\multicolumn{1}{|c|}{\texttt{GPT4}}                       & \multicolumn{1}{c|}{\bf 51.33}    & {\bf 63.66}*    \\ \hline
\multicolumn{1}{|c|}{\texttt{GPT3.5-turbo}}                & \multicolumn{1}{c|}{{49.28}}    & 59.40*    \\ \hline
\multicolumn{1}{|c|}{\texttt{Alpaca-13B} Greedy Search}              & \multicolumn{1}{c|}{37.63}    & 24.06     \\ \hline
\multicolumn{1}{|c|}{\texttt{Alpaca-13B} Beam Search}                & \multicolumn{1}{c|}{40.49}    & 24.31     \\ \hline
\multicolumn{3}{|c|}{\textbf{FANTASE with \texttt{Alpaca-13B} as the Base Model}}                                  \\ \hline
\multicolumn{1}{|c|}{SCD Greedy Search}               & \multicolumn{1}{c|}{42.33}    & 56.64     \\ \hline
\multicolumn{1}{|c|}{SCD Beam Search}                 & \multicolumn{1}{c|}{44.17}    & 62.66     \\ \hline
\multicolumn{3}{|c|}{\cellcolor[HTML]{EFEFEF}{\color[HTML]{000000} \textbf{Supervised Learning Methods}}} \\ \hline
\multicolumn{3}{|c|}{\textbf{Baselines}}                                                                  \\ \hline
\multicolumn{1}{|c|}{\texttt{AlpDSTC-7B} / \texttt{Lynx-7B} Greedy Search}      & \multicolumn{1}{c|}{46.63}     & 48.62     \\ \hline
\multicolumn{1}{|c|}{\texttt{AlpDSTC-7B} / \texttt{Lynx-7B} Beam Search}        & \multicolumn{1}{c|}{47.44}     & 50.53     \\ \hline
\multicolumn{3}{|c|}{\textbf{FANTASE with \texttt{Alpaca-13B} as the Base Model (In-context) and \texttt{RoBERTa-Base} Reranker (Supervised)}}  \\ \hline
\multicolumn{1}{|c|}{Reranking}                               & \multicolumn{1}{c|}{46.42}    & 33.33         \\ \hline
\multicolumn{1}{|c|}{SCD Beam Search + Reranking}     & \multicolumn{1}{c|}{48.88}    & 64.41         \\ \hline
\multicolumn{3}{|c|}{\textbf{FANTASE with \texttt{AlpDSTC-7B} / \texttt{Lynx-7B} as the Base Model}}                          \\ \hline
\multicolumn{1}{|c|}{SCD Greedy Search}               & \multicolumn{1}{c|}{59.30}     & 65.66     \\ \hline
\multicolumn{1}{|c|}{SCD Beam Search}                 & \multicolumn{1}{c|}{{\bf 62.78}}     & {\bf 67.17}     \\ \hline
\end{tabular}
\caption{API Call Generation Accuracy Evaluation On DSTC8 and API Bank. For settings involving beam search, we set the beam size to 4. For reproducible results, we set temperature to 0 for all settings. (* denotes results reported by \citealt{li2023apibank}. Best performed In-context and Supervised Learning Methods are \textbf{bolded}). }
\vspace{-11pt}
\label{table3}
\end{table*}

\section{Experiment Setup}
\subsection{Datasets}
\label{sec:data}
We use the data from \textbf{DSTC8}~\cite{dstc8kim} and \textbf{API Bank}~\cite{li2023apibank} to conduct the experiments and evaluation of our proposed FANTASE framework. Both datasets support the task of API call generation that requires understanding and reasoning of multi-turn human-assistant dialogues. Short-listed APIs and associated Documentation are accompanied by each sample. One major difference is that each DSTC8 sample has two related exemplars while API Bank does not ship with the exemplars. Accordingly, we experiment with the few-shot in-context learning setting with DSTC8 and the zero-shot in-context learning setting with API Bank. In Appendix~\ref{appendix:dataset}, we provide detailed statistics of DSTC8 and API Bank. 

\subsection{Baseline and Backbone Models}
For in-context learning settings, we include strong baselines \texttt{GPT3.5-turbo} and \texttt{GPT4}
developed by OpenAI, which represents the most recent breakthrough in LLMs with a track record of leading zero-shot and few-shot learning and reasoning capabilities. 
FANTASE is a plug-and-play model-agnostic approach that can be used in conjunction with any LLM that has an autoregressive decoder producing next-token probabilities. However, the API access of OpenAI models only allows greedy search and does not provide the logits. In consideration of the license, resource constraints, efficiency, and zero-shot/few-shot learning and reasoning capabilities, we opt to use \texttt{Alpaca-13B} as the backbone of our proposed methods.
As for baselines of supervised learning settings, we fine-tune the \texttt{Alpaca-7B} model with DSTC8 training set samples (denoted as \texttt{AlpDSTC-7B}), and we directly use the \texttt{Lynx-7B} model, an API Bank data tuned \texttt{Alpaca-7B} model, released by \citealt{li2023apibank}. We supply detailed information of the aforementioned models in Appendix~\ref{appendix:models}.

\subsection{Evaluation Settings}
To verify the effectiveness of FANTASE, we run experiments with the following three evaluation settings that focus on different perspectives:

\noindent\textbf{API Call Generation Accuracy} measures if the generated API calls fully match their associated ground truth. It is the main metric that reflects if the generation faithfully followed the user's request and the requirements specified in the API documentation. As the order of arguments does \emph{not} matter for both datasets, we calculate unit-wise order-insensitive set matches. The opposite order-sensitive case is, however, a setting bias in favor of FANTASE as the state-tracking of the SCD component makes it fully aware if it’s going to generate the nth argument. For fair comparison, we follow \citealt{li2023apibank} to report the order-insensitive matching results in this paper.

\noindent\textbf{Inference Efficiency} measures the time cost of the API call generation. Previous works on constrained decoding often vaguely report that the decoding speed is slower than regular decoding algorithms without quantitative measurements. To quantify the speed up brought by FANTASE's State-tracked Constrained Decoding that utilizes CTST (\S\ref{sec:scd}), we compare the time costs of the API call generation with regular/constrained greedy/beam search algorithms using the \texttt{Alpaca-13B} model under the in-context learning setting.

\noindent\textbf{Context Efficiency} measures the effectiveness of our approach in incorporating the API documentation without the reliance on the repeated supply of the lengthy API documentation in the prompt for in-context learning. 

\section{Results and Analysis}
\label{sec:results}
\label{results}
\subsection{API Call Generation Accuracy}
\label{sec:results-gen_acc}
In Table~\ref{table3}, we report the results of API call generation accuracy. 
The SCD component of FANTASE consistently brings substantial improvements over the base models for both in-context learning settings and supervised learning settings on DSTC8 and API Bank, which demonstrates complementary benefits. 

Specifically, for few-shot in-context learning settings evaluated with DSTC8, SCD Greedy Search and SCD Beam Search improve the accuracy by \texttt{+4.7} and \texttt{+3.68} over the respective counterparts. 
For zero-shot in-context learning settings evaluated with API Bank, SCD Greedy Search and SCD Beam Search boost the accuracy by \texttt{+32.33} and \texttt{+34.34} respectively, which makes the \texttt{13B} model's zero-shot generation performance comparable to the \texttt{GPT3.5-turbo} model that has an estimated parameter size of \texttt{175B} and surpasses the accuracy of fine-tuned \texttt{7B} model by a large margin. 
The larger performance gap signifies the value of SCD when labeled data is not available at all. 

For supervised learning settings, SCD can still greatly improve the performance of corresponding settings of fine-tuned models and yields \texttt{+17.04}/\texttt{+16.64} performance gain on API Bank and \texttt{+12.67}/\texttt{+15.34} performance gain on DSTC8 with greedy/beam search, which leads to the \texttt{7B} models outperforming GPT models that are 
much larger in terms of parameter size. 

The Reranking component of FANTASE also brings considerable improvements over the base model by supervised training of a lightweight discriminator. The Reranking component alone improves the regular beam search results by \texttt{+6.15} and \texttt{+9.02} respectively on DSTC8 and API Bank. The FANTASE framework, with both the SCD component and Reranking component activated, achieves the best overall accuracy showing complementary benefits of the two components. Specifically, for DSTC8, the accuracy of \texttt{48.88} is close to the performance of \texttt{GPT3.5-turbo} and better than the supervised fine-tuned model \texttt{AlpDSTC-7B}. For API Bank, the accuracy of \texttt{64.41} is better than \texttt{GPT4} and supervised fine-tuned model \texttt{Lynx-7B}.

\subsection{Inference Efficiency}
\label{sec:results-inf_eff}
\begin{table}[H]
\scriptsize
\centering
\vspace{-11pt}
\begin{tabular}{|c|cc|cc|}
\hline
\multicolumn{1}{|c|}{\multirow{4}{*}{\textbf{\begin{tabular}[c]{@{}c@{}}Decoding \\ Strategy\end{tabular}}}} &
  \multicolumn{2}{c|}{\textbf{DSTC8}} &
  \multicolumn{2}{c|}{\textbf{API Bank}} \\ \cline{2-5} 
\multicolumn{1}{|c|}{} &
  \multicolumn{1}{c|}{\textbf{\begin{tabular}[c]{@{}c@{}}Inference\\ Speed\\(sec/sample)\end{tabular}}} &
  \multicolumn{1}{c|}{\textbf{\begin{tabular}[c]{@{}c@{}}Speed\\Up\end{tabular}}} &
  \multicolumn{1}{c|}{\textbf{\begin{tabular}[c]{@{}c@{}}Inference\\ Speed\\(sec/sample)\end{tabular}}} &
  \multicolumn{1}{c|}{\textbf{\begin{tabular}[c]{@{}c@{}}Speed\\Up\end{tabular}}} \\ \hline
GS & \multicolumn{1}{c|}{5.32}  & -     & \multicolumn{1}{c|}{5.85} & - \\ \hline
SCD GS      & \multicolumn{1}{c|}{3.42}  & x1.56 & \multicolumn{1}{c|}{3.33} & x1.76  \\ \hline
BS   & \multicolumn{1}{c|}{15.12} & -     & \multicolumn{1}{c|}{23.15} & - \\ \hline
SCD BS      & \multicolumn{1}{c|}{6.33}  & x2.39 & \multicolumn{1}{c|}{10.27} & x2.25  \\ \hline
\end{tabular}
\caption{Generation Speed of Regular Greedy Search (GS) and Beam Search (BS) Decoding versus FANTASE's State-tracked Constrained Decoding (SCD) Counterparts.}
\vspace{-11pt}
\label{table4}
\end{table}

In Table~\ref{table4}, we quantitatively measure the inference time savings of the State-Tracked Constrained Decoding that leverages the CTST as we have introduced in Section~\ref{sec:scd} and illustrated in Figure~\ref{figure_ctst}. Compared to regular greedy and beam search, SCD has the capability of speeding up the API call generation by approximately 1.5x\textasciitilde2.4x times. 
% In Figure~\ref{} (TODO - speed vs overall accuracy graph)
When looking together with generation accuracy, it is quite encouraging that SCD greedy search can achieve regular beam search level's performance with significantly less amount of time and SCD Beam Search can achieve much better performance at regular greedy search level's time cost. For API Bank, SCD Greedy search can even outperform regular beam search with significantly less time cost. 

\subsection{Context Efficiency}
\label{sec:results-con_eff}
\begin{table}[H]
\scriptsize
\centering
\vspace{-11pt}
\begin{tabular}{|c|c|c|c|c|}
\hline
\textbf{Dataset} & \textbf{Setting} & \textbf{w. API Doc} & \textbf{w.o. API Doc} & \textbf{$\Delta$} \\ \hline
\multirow{4}{*}{DSTC8}    & GS     & 37.63 & 33.74 & -3.89  \\ \cline{2-5} 
                          & SCD GS & 42.33 & 40.70 & -1.63  \\ \cline{2-5} 
                          & BS     & 40.49 & 38.24 & -2.25  \\ \cline{2-5} 
                          & SCD BS & 44.17 & 42.54 & -1.63  \\ \hline
\multirow{4}{*}{API Bank} & GS     & 24.06 & 4.76  & -19.3  \\ \cline{2-5} 
                          & SCD GS & 56.64 & 22.81 & -33.83 \\ \cline{2-5} 
                          & BS     & 24.31 & 4.51  & -19.8  \\ \cline{2-5} 
                          & SCD BS & 58.65 & 23.05 & -35.6  \\ \hline
\end{tabular}

\caption{Generation Accuracy of Regular Greedy Search (GS) and Beam Search (BS) Decoding with / without API Documentation versus FANTASE's State-tracked Constrained Decoding (SCD) Counterparts.}

\vspace{-11pt}
\label{table5}
\end{table}
As discussed in Section~\ref{sec:scd}, another benefit of SCD is to save the context tokens required in the prompt for supplying the API documentation to the LLM, as SCD is capable of incorporating that information at the decoding stage. Based on the statistics provided in Appendix~\ref{appendix:dataset}, removing API documentation from the input could save an average of 766.28 tokens for DSTC8 and 265.71 tokens for API Bank.
In Table~\ref{table5}, we show the performance of SCD when the API documentation is removed from the model's input.
For DSTC8, our constrained decoding method manages to maintain the accuracy above \texttt{40} when the API Documentation is absent from the input while the unconstrained counterparts suffer larger performance drops of \texttt{-3.89}/\texttt{-2.25} that leads to larger performance gaps with our approach. 
For API Bank, the absence of both exemplars and API documentation makes it super challenging for the model to generate relevant API calls as shown by the large performance drop. However, our constrained decoding method can still achieve \texttt{22.81}/\texttt{23.05} accuracy in such a scenario, which is close to the performance of the unconstrained version that has API Documentation access in the prompt. 

\section{Conclusions}
The FANTASE framework, with its State-Tracked Constrained Decoding (SCD) and Reranking components, effectively tackles the challenges of generating API calls from complex contexts. By integrating API constraints through a Token Search Trie and employing a lightweight model for reranking, FANTASE not only ensures accurate API call generation but also improves inference and context efficiencies. Its superior performance on the DSTC8 and API Bank datasets confirms FANTASE's significant advancement in enhancing large language models' tool-using ability.

\section*{Limitations}
Despite the effort we have made to the best of our current ability, we recognize the following limitations of our work:

\noindent\textbf{Evaluation with larger language models}: Due to our resource limitations and the input length of the two datasets, the largest models we can fine-tune and infer are 7B and 13B respectively. The effect of the base model's parameter size hasn't been extensively studied in this paper, and the impact of our approach on larger models' API call generation accuracy and inference efficiency lacks empirical evidence. Based on the working mechanism of FANTASE, our educated guess is that larger models may make fewer errors which our approach is targeting, so the improvements would be smaller than using relatively small models as the backbone. As for the inference efficiency, the absolute time savings should be larger as a forward pass through a larger model takes longer time. However, the relative speed-up would remain at the current level as the amount of nodes that only have one child depends on the data instead of the models.

\noindent\textbf{Adaptation to research-wise understudied API formats}: As the API call generation datasets available in the research community are based on popular programming languages, especially Python, we implement and verify the effectiveness of FANTASE's SCD component following the conventional API format. Although the high-level idea of SCD is adaptable to research-wise understudied API formats, such adaptation may require moderate engineering effort to replace the structural tokens we used with the respective new format's structural tokens. However, given FANTASE's effectiveness and efficiency, we believe that the benefits would outweigh the adaptation costs especially when the intended use is for large-scale real-world customer-facing applications where accuracy and efficiency matter.

\noindent\textbf{Adaptation to a broader range of tasks}: FANTASE is specially designed and evaluated for the task of our interests - API call generation. Although the high-level idea and concepts of our approach should be adpatable to other structured text generation tasks such as SQL generation, table generation, etc., the adaptation may require considerable efforts in identifying constraints, signature tokens, and appropriate constraint-enforcing steps. For unstructured text generation tasks, it remains unclear if the identification of signature tokens and appropriate constraint-enforcing steps are feasible.

\section*{Ethics Statement}
Our approach significantly enhances the capability of current models to generate API calls. However, it's important to acknowledge that the accuracy of these generations remains imperfect. As API calls could enable language models to perform tangible real-world actions, inaccuracies in API call generation would lead to serious consequences. These may include but are not limited to: financial losses when making wrong purchases and reservations, potential harm to the human being or the environment when controlling physical objects in unexpected ways, and the dissemination of false or misleading information when retrieving a wrong set of information. Consequently, we urge the users of our methods and the related API call generation models to be aware of such systems' high likelihood of generating inaccurate API calls and to remain vigilant about the possible risks associated with such errors.

\bibliography{anthology,custom}
\bibliographystyle{acl_natbib}

% \clearpage
\appendix

\section*{\centering Appendix}

\section{Datasets}
\label{appendix:dataset}

\begin{table}[H]
\scriptsize
\centering

\begin{tabular}{|c|c|c|}
\hline
\textbf{Dataset} &
  \begin{tabular}[c]{@{}c@{}}\textbf{DSTC8}\\ median / mean\end{tabular} &
  \begin{tabular}[c]{@{}c@{}}\textbf{API Bank}\\ median / mean\end{tabular} \\ \hline
\textbf{\begin{tabular}[c]{@{}c@{}}Total Input \\ Length (Tokens)\end{tabular}} &
  \begin{tabular}[c]{@{}c@{}}1,683\\ 1,644.28\end{tabular} &
  \begin{tabular}[c]{@{}c@{}}492\\ 542.86\end{tabular} \\ \hline
\textbf{\begin{tabular}[c]{@{}c@{}}API Documentation\\ Length (Tokens)\end{tabular}} &
  \begin{tabular}[c]{@{}c@{}}735\\ 766.28\end{tabular} &
  \begin{tabular}[c]{@{}c@{}}244\\ 265.71\end{tabular} \\ \hline
\textbf{\begin{tabular}[c]{@{}c@{}}Exemplars\\ Length (Tokens)\end{tabular}} &
  \begin{tabular}[c]{@{}c@{}}580\\ 558.00\end{tabular} &
  N/A \\ \hline
\textbf{\begin{tabular}[c]{@{}c@{}}Conversation\\ Length (Tokens)\end{tabular}} &
  \begin{tabular}[c]{@{}c@{}}200.5\\ 221.08\end{tabular} &
  \begin{tabular}[c]{@{}c@{}}138\\ 175.15\end{tabular} \\ \hline
\textbf{\begin{tabular}[c]{@{}c@{}}Conversation\\ Turns\end{tabular}} &
  \begin{tabular}[c]{@{}c@{}}6\\ 6.31\end{tabular} &
  \begin{tabular}[c]{@{}c@{}}3\\ 3.21\end{tabular} \\ \hline
\textbf{\begin{tabular}[c]{@{}c@{}}Target API Call\\ Length (Tokens)\end{tabular}} &
  \begin{tabular}[c]{@{}c@{}}46\\ 45.70\end{tabular} &
  \begin{tabular}[c]{@{}c@{}}27\\ 35.17\end{tabular} \\ \hline
\textbf{\begin{tabular}[c]{@{}c@{}}Target API Call\\ Arguments Amount\end{tabular}} &
  \begin{tabular}[c]{@{}c@{}}3\\ 3.42\end{tabular} &
  \begin{tabular}[c]{@{}c@{}}2\\ 2.27\end{tabular} \\ \hline
\end{tabular}

\caption{Statistics of DSTC8 and API Bank Data.}

\label{table6}
\end{table}
In Table~\ref{table6}, we provide the detailed statistics of DSTC8 and API Bank. The inputs and outputs of DSTC8 are longer than API Bank counterparts. DSTC8 also has more turns of conversation and API call arguments than API Bank, which makes DSTC8 a more challenging dataset in terms of reasoning the context and generating appropriate API calls even if two exemplars are provided for each sample. 

Our experiments and evaluation of the FANTASE framework are based on the test split of DSTC8 and API Bank that has 490 and 399 samples respectively. For DSTC8, we remove the longest one out of the 490 samples as it causes CUDA out of memory error on our server. 

\section{Models}
\label{appendix:models}
\noindent\textbf{\texttt{Alpaca-13B}} is a \texttt{LLaMA}-based instruction following model that has 40 layers, a hidden size of 5120, 40 self-attention heads and 13 billion parameters. For our experiments, we use Huggingface checkpoint \texttt{chavinlo/gpt4-x-alpaca}. 

\noindent\textbf{\texttt{AlpDSTC-7B} and \texttt{Lynx-7B}} are the \texttt{Alpaca-7B}-based model that has 32 layers, a hidden size of 4096, 32 self-attention heads and 7 billion parameters. \texttt{Lynx-7B} is the training set fine-tuned model released by API Bank authors (Huggingface checkpoint: \texttt{liminghao1630/Lynx-7b}). The model was tuned for 3 epochs with a learning rate of 2e-5 and an effective batch size of 256. We follow the same setting of \texttt{Lynx-7B} to tune a \texttt{Alpaca-7B} with DSTC8 training set samples. 

\noindent\textbf{\texttt{RoBERTa-base}} is a \texttt{BERT}-based model that has 12 layers, a hidden size of 768, 12 self-attention heads, and 125 million parameters. For our experiments, we use Huggingface checkpoint \texttt{roberta-base} and tune the model with training set candiate generations for 5 epochs with a learning rate of 5e-5 and an effective batch size of 256. 

\noindent\textbf{\texttt{GPT3.5} and \texttt{GPT4}} are only accessible via API or web interface. Details of these two models have not been officially released by OpenAI, but a broadly accepted latency-based parameter size estimation is 175 billion parameters for \texttt{GPT3.5} and 1.76 trillion parameters for \texttt{GPT4}. For our experiments, we use checkpoints \texttt{gpt-3.5-turbo-0613} and \texttt{gpt-4o-2024-05-13}. 

\noindent License information of \texttt{Alpaca} can be found at \url{https://github.com/tatsu-lab/stanford_alpaca/blob/main/LICENSE} and \url{https://github.com/tatsu-lab/stanford_alpaca/blob/main/DATA_LICENSE}. 

\noindent License information of \texttt{LLaMA} can be found at \url{https://docs.google.com/forms/d/e/1FAIpQLSfqNECQnMkycAp2jP4Z9TFX0cGR4uf7b_fBxjY_OjhJILlKGA/viewform}.

\noindent License information of \texttt{Lynx-7B} and API Bank data can be found at \url{https://github.com/AlibabaResearch/DAMO-ConvAI/blob/main/api-bank/LICENSE}. 

\noindent License information of DSTC8 data can be found at \url{https://github.com/google-research-datasets/dstc8-schema-guided-dialogue/blob/master/LICENSE.txt}.

\noindent License information of \texttt{RoBERTa} can be found at \url{https://github.com/facebookresearch/fairseq/blob/main/LICENSE}.

\noindent License information of \texttt{GPT-X} models can be found at \url{https://openai.com/policies/terms-of-use}.

\section{Comparative Summarization}
In Table~\ref{comp}, we summarize the pros and cons of different methods based on the experiments and analysis we have reported in this paper for a more nuanced understanding of FANTASE's strengths and weaknesses in relation to other approaches. 

\begin{table*}[]
\scriptsize
\centering
\begin{tabular}{|c|c|c|c|c|c|c|}
\hline
\textbf{Method} & \textbf{\begin{tabular}[c]{@{}c@{}}Labeled Data\\ Amount\end{tabular}} & \textbf{\begin{tabular}[c]{@{}c@{}}Learning\\ Efficiency\end{tabular}} & \textbf{\begin{tabular}[c]{@{}c@{}}Contex\\ Requirement\end{tabular}} & \textbf{\begin{tabular}[c]{@{}c@{}}Faithfulness to\\ API Documentation\end{tabular}} & \textbf{\begin{tabular}[c]{@{}c@{}}Generation\\ Accuracy\end{tabular}} & \textbf{\begin{tabular}[c]{@{}c@{}}Generation\\ Efficiency\end{tabular}} \\ \hline
\begin{tabular}[c]{@{}c@{}}Fine-tuning\\ with Regular Decoding\end{tabular} & Large & Worst & Low & \begin{tabular}[c]{@{}c@{}}Not\\ Guaranteed\end{tabular} & Better & Worst \\ \hline
\begin{tabular}[c]{@{}c@{}}Fine-tuning\\ with FANTASE-SCD\end{tabular} & Large & Worst & Low & Guaranteed & Best & Best \\ \hline
\begin{tabular}[c]{@{}c@{}}In-context Learning\\ with Regular Decoding**\end{tabular} & Small or Zero & Best & High & \begin{tabular}[c]{@{}c@{}}Not\\ Guaranteed\end{tabular} & Worst & Worst \\ \hline
\begin{tabular}[c]{@{}c@{}}In-context Learning\\ with FANTASE-SCD\end{tabular} & Small or Zero & Best & Moderate & Guaranteed & Better & Best \\ \hline
\begin{tabular}[c]{@{}c@{}}In-context Learning\\ with FANTASE-SCD and\\ Fine-tuned Reranker\end{tabular} & Large & Better & Moderate & Guaranteed & Much Better & Better \\ \hline
\end{tabular}
\caption{Comparative Summarization \scriptsize(** Summarized from the results and analysis of the 13B model for fair comparison)}
\label{comp}
\end{table*}

\end{document}